\pdfoutput=1

\documentclass[10pt,twocolumn,letterpaper]{article}

\usepackage{times}
\usepackage{epsfig}
\usepackage{graphicx}
\usepackage{amsmath}
\usepackage{amssymb}
\usepackage{caption}
\usepackage{subcaption}
\usepackage{multirow}

\usepackage[pagebackref=true,breaklinks=true,colorlinks,bookmarks=false]{hyperref}

\newcommand{\eg}{e.g.\ }
\newcommand{\ie}{i.e.\ }
\newcommand{\etal}{\textit{et al.\ }}

\begin{document}

\title{Neural Human Deformation Transfer}

\date{}

\author{Jean Basset \thanks{Univ. Grenoble Alpes, Inria, CNRS, Grenoble INP (Institute of Engineering Univ. Grenoble Alpes), LJK, 38000 Grenoble, France} \\
{\tt\small jean.basset@inria.fr}
\and
Adnane Boukhayma \thanks{Univ Rennes, Inria, CNRS IRISA, M2S, France} \\
{\tt\small adnane.boukhayma@inria.fr}

\and
Stefanie Wuhrer \footnotemark[1]\\
{\tt\small stefanie.wuhrer@inria.fr}

\and
Franck Multon \footnotemark[2]\\
{\tt\small fmulton@irisa.fr}

\and
Edmond Boyer \footnotemark[1]\\
{\tt\small edmond.boyer@inria.fr}
}


\maketitle

\thispagestyle{empty}

\begin{abstract}
We consider the problem of human deformation transfer, where the goal is to retarget poses between different characters. Traditional methods that tackle this problem assume a human pose model to be available and transfer poses between characters using this model. In this work, we take a different approach and transform the identity of a character into a new identity without modifying the character's pose. This offers the advantage of not having to define equivalences between 3D human poses, which is not straightforward as poses tend to change depending on the identity of the character performing them, and as their meaning is highly contextual. To achieve the deformation transfer, we propose a neural encoder-decoder architecture where only identity information is encoded and where the decoder is conditioned on the pose. We use pose independent representations, such as isometry-invariant shape characteristics, to represent identity features. Our model uses these features to supervise the prediction of offsets from the deformed pose to the result of the transfer. We show experimentally that our method outperforms state-of-the-art methods both quantitatively and qualitatively, and generalises better to poses not seen during training. We also introduce a fine-tuning step that allows to obtain competitive results for extreme identities, and allows to transfer simple clothing.
\end{abstract}

\section{Introduction}



Deformation transfer is the process of retargeting poses between characters: Given a source character in a deformed pose, and a target character in a reference pose, the objective is to generate a new shape of the target character in the deformed pose~\eg~\cite{sumner2004deformation}. It finds interest in digital content creation where it has the potential to drastically reduce animation costs. This is particularly relevant when applied to human shapes, where it can ease animation production from captured motions and enable 3D and 4D data augmentation for data-driven applications. 

Deformation transfer between humans requires poses to be defined consistently across different characters. This relies on assumptions on human identities and poses, which are inherently entangled notions. Human poses can be assumed to be identifiable in a coherent way over different characters. In that case, deformation transfer boils down to transferring the given deformed pose to the target character, and the result can thus be seen as the target identity with a new pose. Such a transfer can be achieved through a shared continuous pose parameterisation~\eg~\cite{anguelov2005scape,loper2015smpl} or through a discrete correspondence map, as with style transfer~\eg~\cite{marin2020instant}.
Another approach is to assume that the human shape can be characterised independently of the pose,~\ie by its identity. In this case, retargeting can be performed by transferring the identity from the target to the source characters, and the result can be seen as the source deformed pose with a new identity.

Both interpretations of the deformation transfer problem are arguably approximations since exact correspondences between character poses are subjective and since human shapes are not fully independent of the pose. They anyway enable practical solutions as illustrated in the literature. 
While numerous methods have been proposed to solve the pose transfer problem (\eg~\cite{sumner2004deformation,gleicher1998retargetting,ho2010spatial,kulpa2005morphology,baran2009semantic}), few works address the alternative solution with identity transfer~\cite{basset2020contact}. However the latter exhibits advantages, in particular the ability to better adapt to any pose by directly considering the correct pose and just modifying identity shape properties.

In this paper we investigate the identity transfer strategy with a data-driven approach. We propose a deep learning architecture that predicts the deformation of the source model so that its identity matches that of the target model. The architecture of the model consists of an encoder that encodes the identity of the target model into a low-dimensional feature vector, and a decoder that consumes the identity feature vector along with the source model and predicts offsets from the source model that transfer the identity. To encode identity information, losses based on classical assumptions of human deformations are used, namely losses based on the hypotheses that two models sharing the same identity should be near-isometric~\cite{cosmo2019isospectralization} and have body parts that deform near-rigidly between the poses~\cite{sorkine2007rigid}. To structure the latent space, we use a loss that aims to map feature vectors of the same identity to the same location in latent space.

We train our architecture in a weakly supervised way: while we rely on the presence of identity labels for all training data, we only require pose labels for a small subset since 3D models with different characters performing the exact same pose are rare in existing datasets. To have access to high-quality labelled data, we propose an extension of the FAUST dataset~\cite{bogo2014FAUST} that includes additional poses and identities with full label information. We demonstrate experimentally that having access to full label information, and hence a reconstruction loss, on a small proportion of the training data is sufficient to train our architecture.

Inspired by the few-shot learning of generative models literature (\eg~\cite{zakharov2019few,arik2018neural,jia2018transfer}), we observe that fine-tuning our feed forward network at test time improves the results. This is achieved with a few extra training steps on the inputs using a self-supervised loss. In this strategy, the initial network training can be seen as a meta-learning stage, and the fine-tuning can be interpreted as one-shot learning from a single reference pose / target identity pair, which adapts the network weights further to that specific case. To the best of our knowledge, this is the first learning-based deformation transfer work that explores such an idea. Not only does the fine-tuning improve our performance quantitatively, but it also allows us to successfully transfer identity for out of training distribution shapes, such as a shape of a simply clothed person with a hat and a backpack, while the training consists merely of minimally dressed body shapes.   

We compare our method to deformation transfer results by the recent deep learning approaches Unsupervised Shape and Pose Disentanglement (USPD)~\cite{zhou20unsupervised} and Neural Pose Transfer (NPT)~\cite{wang2020neural}, and show that geometric detail is better preserved with our method when applied to poses not observed during training.

In summary, our contributions are:
\begin{itemize}
    \item our method better generalises to poses not seen during training than state-of-the-art, achieved by transferring the identity of the target shape to the deformed pose within a deep learning framework.
    \item our method allows to preserve fine-scale detail linked to the identity of the character and generalizes to characters wearing simple clothing thanks to test time identity transfer refinement with fine-tuning.
    \item we extend the FAUST dataset~\cite{bogo2014FAUST} to contain more identities and poses with full label information, which can be leveraged for training.
\end{itemize}

\section{Related Work}

This section reviews works that solve the deformation transfer problem, generative models for deep learning and recent deep-learning methods for deformation transfer.

\textbf{Deformation transfer}. Deformation transfer aims to deform a character to make it mimic the pose of a source character. This is classically done either by adapting the skeleton movement to a new skeleton~\cite{gleicher1998retargetting, lee1999hierarchical, kulpa2005morphology, ho2010spatial}, or by directly deforming the surface of the characters~\cite{sumner2004deformation, zhou2010deformation, baran2009semantic, liu2018surface}.
Recently, Basset~\etal~\cite{basset2020contact} showed that transferring the shape, while preserving the pose, requires simpler deformations than deforming the pose of the target character. We follow a similar direction by predicting the shape deformation from the source to the result, instead of predicting coordinates from scratch.
These methods can give good results, but often come at an important computational cost. Instead of using an optimisation based technique, we thus chose to leverage recent advances in deep learning.

\textbf{Generative deep learning models}. Deep learning methods have demonstrated a strong generative capability. Recent works have focused on learning to generate 3D data, in particular with autoencoders. Jiang~\etal~\cite{jiang2020disentangled} represent 3D data by encoding the vertices of their meshes in a lower dimensional feature, based on an anatomical hierarchical segmentation, and train their model on those features. We make use of similar information by segmenting the meshes into body parts. Similarly, Tan~\etal~\cite{tan2018variational} represent meshes with rotation invariant features and train a variational encoder with fully connected layers on these features. Ranjan~\etal~\cite{ranjan2018generating} define a spectral convolutional layer to generate 3D human faces. They also introduce down and up sampling layers adapted to 3D mesh data based on quadratic edge collapse. Bouritsas~\etal~\cite{bouritsas2019neural} use a spiral ordering of the neighbourhood of each vertex to apply convolutions to 3D meshes. These methods give satisfying results for generating 3D data, and  our architecture is based on the building blocks they provide.

Tretschk~\etal~\cite{tretschk2020demea} predict the rigid deformation from a template mesh to their output, which results in higher quality reconstructions of poses. We use a similar idea but take it a step further; instead of predicting the deformation from a common template, we predict the deformation directly from the pose we want to preserve.

\textbf{Deep learning for deformation transfer}.
Recently, deep learning methods have been applied to the deformation transfer problem, on various categories of inputs. Numerous works explored transferring 2D video between animated characters~\eg~\cite{chan2019everybody,villegas2018neural} or videos of real people~\eg~\cite{liu2019liquid,esser2018variational,kappel2021high}. Some works explored transferring the style between rigid 3D objects. For instance, Wang~\etal~\cite{wang20193dn}  propose to transfer the style of 3D objects by conditioning their decoder on the vertex coordinates of the object to be deformed, and predicting offsets from this source to the result. This is close to the shape transfer idea by Basset~\etal~\cite{basset2020contact}, and we use a similar architecture.

Gao~\etal~\cite{gao18unsupervised} trained two autoencoders for the source and the target shapes, and used a GAN to map the latent code of a deformed source to the latent code of the deformed target. This leads to satisfying deformation transfer results, but the model needs to be retrained for each new shape pair.

Closer to our method, some works~\cite{marin2020instant, cosmo2019isospectralization} describe the instrinsic shape, or style, of a 3D model using the spectrum of its Laplace Beltrami operator (LBO). This is made possible by the observation that the LBO spectrum is invariant to isometric deformation. Style transfer is performed by aligning the spectrum of the source pose to the one of the style target. However the LBO spectrum is known to be sensitive to noise, and struggles to encode fine details of the shape, which limits the transfer results to simple shape classes or to only smoothed versions of the target.

Recently a lot of interest has been given to autoencoders to disentangle shape and pose parameters~\cite{cosmo20limp,zhou20unsupervised,jiang2020disentangled,aumentado2019geometric}. These methods naturally allow deformation transfer by reconstructing a character from the shape and pose parameters of two different characters. A common problem these methods encounter is the lack of large real world datasets with pose labels. 
To remedy this, Cosmo~\etal~\cite{cosmo20limp} present LIMP, a supervised model that allows to train from small-scale datasets. LIMP is built on the hypothesis that human pose deformations are near-isometric, and preserve geodesic distances. The computational cost of the metrics makes LIMP unscalable to large datasets. 
Another supervised deformation transfer method uses spatially adaptive instance normalisation to perform pose transfer between human characters~\cite{wang2020neural}. While this method, called Neural Pose Transfer (NPT), has access to identity and pose labels for training, the training 3D models do not need to be in correspondence, unlike  other methods.
Another approach to address the data problem is to train in an unsupervised manner. Aumentado-Armstrong~\etal~\cite{aumentado2019geometric,aumentado2021disentangling} use the LBO spectrum to define intrinsic shapes in a way invariant to isometric pose deformation. This allows for unsupervised decoupling of shapes and poses, but the method is sensible to the aforementioned limitations of the LBO spectrum. Zhou~\etal~\cite{zhou20unsupervised} create pseudo ground truth for the pose transfer between two characters, by applying on the fly  As Rigid As Possible deformations~\cite{sorkine2007rigid} during training. This method, Unsupervised Shape and Pose Disentanglement (USPD), constitutes the state-of-the-art for unsupervised methods.

\section{Neural Identity Transfer}

This section describes our neural identity transfer method. 
We address the problem of deformation transfer between 3D shapes described by triangle meshes with the same topology,~\ie all meshes have the same connectivity and vertex to vertex correspondence. We assume a dataset of such shapes~\ie meshes $\{\mathcal{M}\}$ where some have ground-truth identity and/or pose labels, and denote them by $\mathcal{M}^{id}_{p}$, $id$ being the identity label and $p$ the pose label.   


\subsection{Overview}

\begin{figure}[t]
\includegraphics[width=\linewidth]{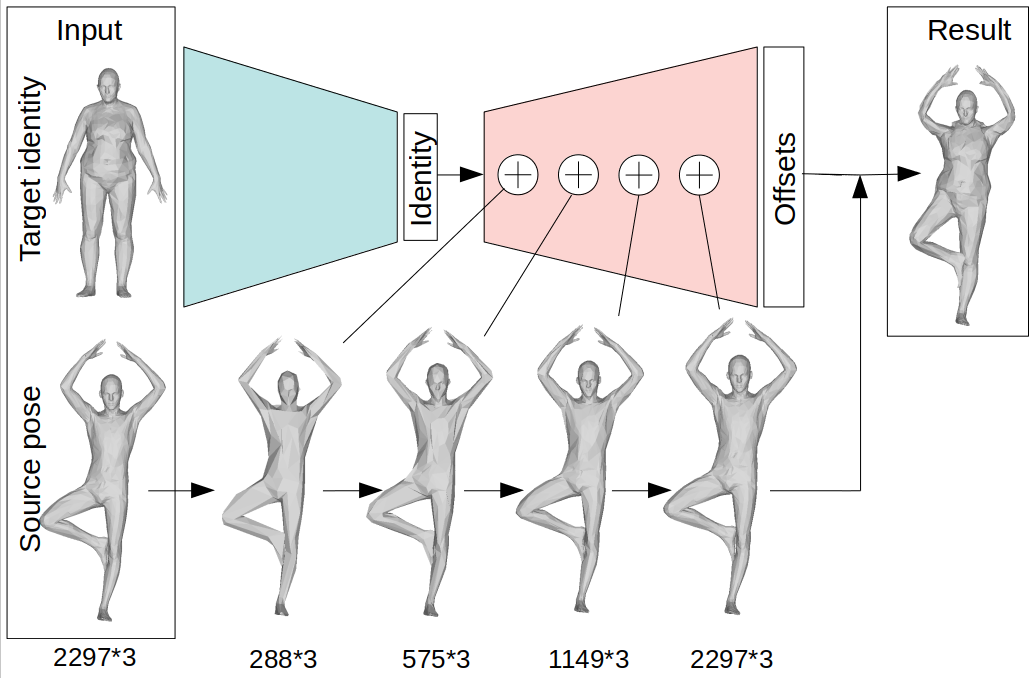}
\centering
\caption{Overview of the proposed approach. The encoder (green) generates an identity code for the target. We feed this code to the decoder (red) along with the source, which is concatenated with the decoder features at all resolution stages. The decoder finally outputs per vertex offsets from the input source towards the identity transfer result.   
}
\label{fig:pipe}
\end{figure}

Fig.~\ref{fig:pipe} provides a visual overview of our approach. Given two meshes, $\mathcal{M}_{p}$ with an input source pose and $\mathcal{M}^{id}$ with an input target identity, our goal is to generate a third mesh $\mathcal{\tilde{M}}^{id}_{p}$ representing the shape of the target identity $id$ in the source pose $p$. We formulate this problem using a deep learning framework and an encoder-decoder architecture. 

Our neural architecture implements the identity transfer by predicting the deformation of the source model $\mathcal{M}_{p}$ so that its identity matches that of the target model $\mathcal{M}^{id}$. This does not require to encode explicitly pose information as the pose is naturally preserved by predicting identity deformations only. 

The encoder $Enc$ takes $\mathcal{M}^{id}$ as input and encodes its identity information into a low-dimensional feature vector $\boldsymbol{z}^{id}$ as 
\begin{equation}
\boldsymbol{z}^{id} = Enc(\mathcal{M}^{id}).
\end{equation}
The decoder $Dec$ takes as input a latent code $\boldsymbol{z}^{id}$ along with $\mathcal{M}_{p}$ and outputs offsets from $\mathcal{M}_{p}$ to $\mathcal{\tilde{M}}^{id}_{p}$ as
\begin{equation}
\label{eq:transfer}
\mathcal{\tilde{M}}_p^{id} = Dec(\boldsymbol{z}^{id},\mathcal{M}_p) + \mathcal{M}_p.
\end{equation}
The architectures of both $Enc$ and $Dec$ are based on spiral convolutions at gradually decreasing/increasing mesh resolutions through pooling/unpooling layers~\cite{bouritsas2019neural}. Note that if the pose information is not explicitly encoded,  $Dec$ is anyway conditioned on the pose $\mathcal{M}_{p}$. This is achieved in practice by concatenating channel-wise at every convolution and unpooling layer of $Dec$ the current vertex features and the 3D coordinates of $\mathcal{M}_{p}$ at the corresponding mesh resolution.

The two main differences in terms of architecture compared to state-of-the-art deformation transfer methods, such as~\cite{cosmo20limp,zhou20unsupervised}, result from the identity transfer strategy. First, rather than encoding pose information, our decoder is conditioned on the input model $\mathcal{M}_{p}$ that has the desired pose. Second, rather than predicting 3D vertex information directly, our decoder predicts offsets from $\mathcal{M}_{p}$. 

At inference time, we fine-tune our feed-forward network to improve the results. This strategy, where network training can be seen as a meta-learning stage and fine-tuning as one-shot learning from a single $\mathcal{M}_{p},\mathcal{M}^{id}$ pair, is inspired by the few-shots learning of generative models literature (\eg~\cite{zakharov2019few,arik2018neural,jia2018transfer}).

\subsection{Training}
\label{sec:learningobjective}

The model is trained in a weakly supervised way because labeled 3D models of different characters performing the exact same poses are rare in existing datasets. In particular, while each model is equipped with an identity label, only a small subset of all models is equipped with a pose label. For training, we sample triplets of distinct meshes of the form $(\mathcal{M}_{p_1}^{id_1},\mathcal{M}_{p_2}^{id_2},\mathcal{M}_{p_1}^{id_2})$ for fully labeled data, and of the form $(\mathcal{M}_{p_1}^{id_1},\mathcal{M}_{p_2}^{id_2},\mathcal{M}_{p_3}^{id_2})$ for data with only identity labels (with unknown pose labels $p_1,p_2,p_3$). Note that while fully labeled data contains the ground truth of the deformation transfer result $\mathcal{M}_{p_1}^{id_2}$, this information is not available for data with identity labels only.

These triplets are used to train the network based on the following losses
\begin{eqnarray}
    l_{sup} &=& \alpha_{lat} l_{lat} + \alpha_{rec} l_{rec}, \nonumber\\
    l_{weaksup} &=& \alpha_{lat} l_{lat} + \alpha_{lap} l_{lap} + \alpha_{rig} l_{rig},
\label{equ:loss}    
\end{eqnarray}
where $l_{sup}$ is the supervised loss used when full label information is available and $l_{weaksup}$ is the weakly supervised loss used when merely identity labels are known. 
Let $\mathcal{\tilde{M}}_{p_1}^{id_2}$ denote the transfer result predicted by our method for inputs $\mathcal{M}_{p_1}^{id_1}$ as source pose and $\mathcal{M}_{p_2}^{id_2}$ as target identity.

We use three types of losses to train the network. First, a latent loss $ l_{lat}$, which helps structuring the latent space, is used during both full and weak supervision. Second, in case of full supervision, a standard $L_2$ penalty reconstruction loss $l_{rec}$ is employed. Finally, in case of weak supervision, two self supervised identity losses $l_{lap}$ and $l_{rig}$ are used that measure identity distances  based on pre-defined deformation models. These losses are weighted using the weights $\alpha_{lat}$, $\alpha_{rec}$, $\alpha_{lap}$, and $\alpha_{rig}$. Details on these losses follow.   

\paragraph{Latent loss}

This loss  uses of the identity label of our data, and constrains the identity latent space by encouraging pairs of shapes that share the same identity to have similar latent representations as
\begin{equation}
    l_{lat}(\mathcal{M}_{p_1}^{id_1},\mathcal{M}_{p_2}^{id_1}) = \|Enc(\mathcal{M}_{p_1}^{id_1}) - Enc(\mathcal{M}_{p_2}^{id_1})\|^2_2.
\end{equation}
This loss is evaluated for the two input meshes of the triplet that share the same identity code, namely $\mathcal{M}_{p_2}^{id_2},\mathcal{M}_{p_1}^{id_2}$ for fully labeled data and $\mathcal{M}_{p_2}^{id_2},\mathcal{M}_{p_3}^{id_2}$ for data with identity labels only.

\paragraph{Reconstruction Loss}

When pose labels are available, we use a standard reconstruction loss that measures the vertex-to-vertex $L_2$ distance between the ground truth $\mathcal{M}_{p_1}^{id_2}$ and the predicted result $\mathcal{\tilde{M}}_{p_1}^{id_2}$ as
\begin{equation}
    l_{rec}(\mathcal{\tilde{M}}_{p_1}^{id_2},\mathcal{M}_{p_1}^{id_2}) = \|\mathcal{\tilde{M}}_{p_1}^{id_2}-\mathcal{M}_{p_1}^{id_2}\|^2_2.
\end{equation}
This strong constraint, that is effective for training, is only required for a small subset of our training data.

\paragraph{Identity Losses} 

\begin{figure}[t]
    \centering
    \begin{subfigure}{0.9\linewidth}
        \includegraphics[width=\linewidth]{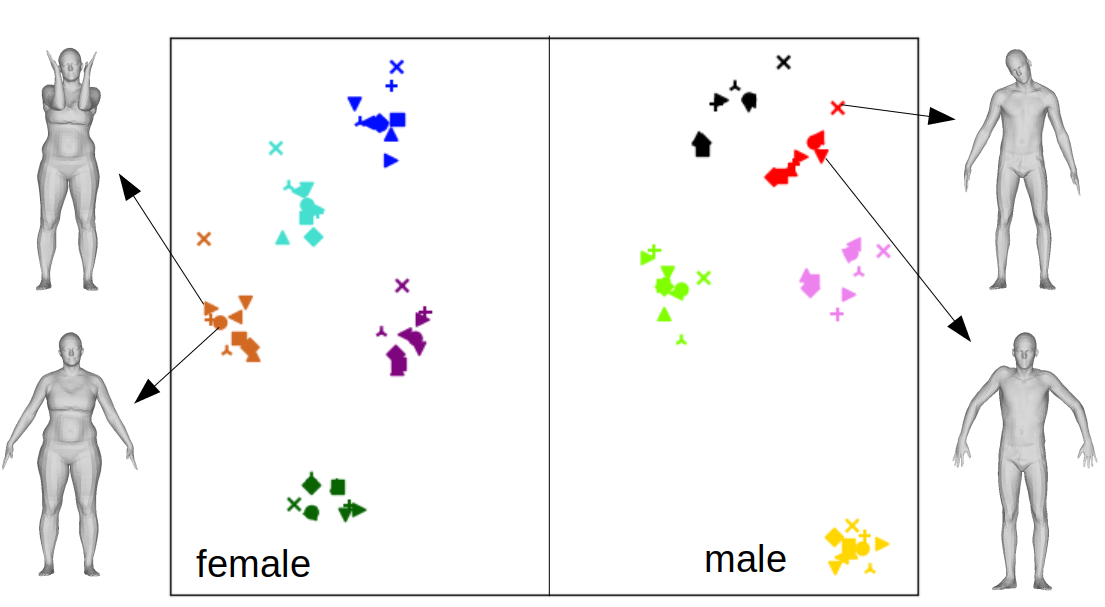}
        \caption{Local laplacian coordinates}
        \label{fig:tsne_iso}
    \end{subfigure}
    
    \begin{subfigure}{0.9\linewidth}
        \includegraphics[width=\linewidth]{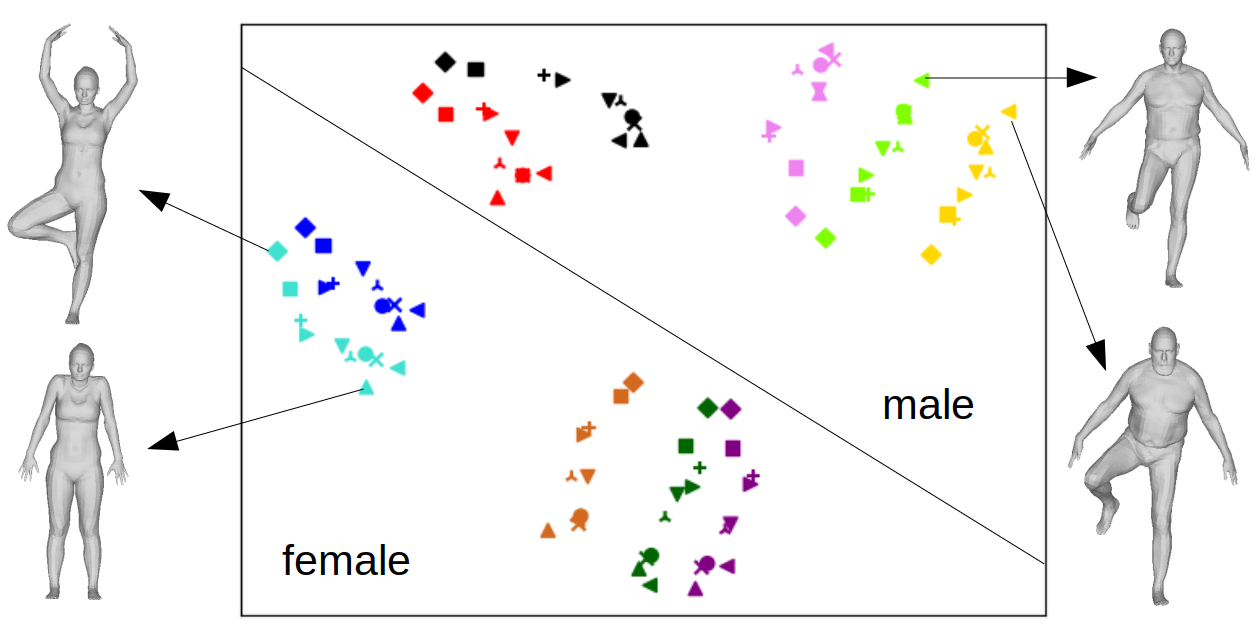}
        \caption{Distances between vertices belonging to the same body part}
        \label{fig:tsne_intrabp}
    \end{subfigure}
    
    \caption{T-SNE dimensionality reduction applied to local Laplacian~(\ref{fig:tsne_iso}) and intra body part distances~(\ref{fig:tsne_intrabp}). All parameters are computed on the FAUST dataset, containing 10 identities performing 10 poses. In each figure, marker colors indicate identities, and marker shapes indicate poses.}
    \label{fig:tsne}
\end{figure}

When only identity labels are available, we evaluate differences in identity using two common hypotheses on human deformations. 

The first hypothesis is that two characters with the same identity are near-isometric~\cite{cosmo2019isospectralization}. We use in particular an unsupervised loss that measures differences of the local Laplacian coordinates~\cite{wuhrer2012posture} between our prediction and the target identity. We validate the suitability of this loss using the T-SNE dimensionality reduction method~\cite{van2008visualizing} on the FAUST dataset~\cite{bogo2014FAUST} that contains 10 identities in 10 different poses each. Fig.~\ref{fig:tsne_iso} shows the result, which clearly groups models with the same identity. This validates our hypothesis that near-isometric meshes, approximated with local Laplacian coordinates, are likely to represent the same identity. To formally define our loss, let $L$ be the Laplacian matrix with uniform weights associated with our common mesh connectivity. We compute the Laplacian coordinates of a mesh as $\Delta = L\mathcal{M}$. We then make these coordinates pose invariant by expressing them in per-vertex local coordinate frames, and denote the result by $\Delta_{loc}$. Meshes that preserve the resulting local coordinates are near-isometric. Our loss between two meshes $\mathcal{M}_{p_1}^{id}$ and $\mathcal{M}_{p_2}^{id}$ of the same identity is
\begin{equation}
    l_{lap}(\mathcal{M}_{p_1}^{id},\mathcal{M}_{p_2}^{id}) = \| {\Delta_{loc}}_{p_1}^{id} - {\Delta_{loc}}_{p_2}^{id} \|^2_2.
\end{equation}
We use this loss between our prediction $\mathcal{\tilde{M}}_{p_1}^{id_2}$ and the given model $\mathcal{M}_{p_2}^{id_2}$ during training.

The second hypothesis is that body parts of a same identity deform near-rigidly between different poses. To validate this hypothesis, Fig.~\ref{fig:tsne_intrabp} shows a T-SNE plot for FAUST data between all Euclidean intra body part distances. Note that identities are well clustered, which validates our hypothesis. Our loss penalizes distances between vertices belonging to the same body part being inconsistent between our prediction and the target identity. The unsupervised rigidity loss between two models $\mathcal{M}_{p_1}^{id}$ and $\mathcal{M}_{p_2}^{id}$ of the same identity is then

\begin{multline}
    l_{rig}(\mathcal{M}_{p_1}^{id},\mathcal{M}_{p_2}^{id})\\
    = \sum\limits_{P \in \mathcal{P}} \sum\limits_{i,j \in P} \|d(\boldsymbol{v}_{i,1}, \boldsymbol{v}_{j,1}) - d(\boldsymbol{v}_{i,2},\boldsymbol{v}_{j,2})\|^2,
\end{multline}

where $\mathcal{P}$ is the set of mesh body parts, $\{\boldsymbol{v}_{i,k}\}$ are the vertices of $\mathcal{M}_{p_k}^{id}$ and $d(.,.)$ is the Euclidean distance. We use this loss between our prediction $\mathcal{\tilde{M}}_{p_1}^{id_2}$ and the given model $\mathcal{M}_{p_2}^{id_2}$ during training.

Note that the two identity losses are evaluated between our prediction and the target identity used for the prediction, and are thus fully unsupervised.

\subsection{Fine-Tuning}
\label{sec:finetuning}


We introduce a fine-tuning step that is performed systematically at test time. At a small additional computational cost, this step allows  to improve results, and enables identity transfer to new shapes considerably different from those seen during training, as demonstrated experimentally.

This step acts as an additional adaptation of the weights of our pre-trained network to a specific input. Given a target identity $\mathcal{M}^{id}$ and a source pose $\mathcal{M}_{p}$, we first generate our result $\tilde{\mathcal{M}}^{id}_{p}$ using the trained model as described in Eq.~\ref{eq:transfer}. This result is used as initialisation for further optimisation. We fine-tune our model for a few more iterations, using as input identity the target identity $\mathcal{M}^{id}$, and as input pose the initial inference result $\tilde{\mathcal{M}}^{id}_{p}$. For these extra training steps, we use a self-supervised loss, combining the Laplacian and the rigidity losses to maintain the target identity, in addition to a regularization loss $l_{reg}$ in the form of 
a $L_2$ penalty between the vertices of the initial result $\tilde{\mathcal{M}}^{id}_{p}$ and those of the final fine-tuned mesh
\begin{equation}
    l_{ft} = \alpha_{lap} l_{lap} + \alpha_{rig} l_{rig} + \alpha_{reg} l_{reg}.
    \label{equ:loss_ft}
\end{equation}






\subsection{Implementation Details}
\label{sec:implementation}

For the main model training, we generate training triplets as follows: a triplet $(\mathcal{M}_1,\mathcal{M}_2,\mathcal{M}_3)$ is created for each mesh sample $\mathcal{M}_1$ in the training data. The third and second meshes of the triplet $\mathcal{M}_3$ and $\mathcal{M}_2$ are then randomly sampled from the same identity as the first mesh, and other identities, respectively. If $\mathcal{M}_1$ comes from the portion of our data with pose labels, we restrict the choice of $\mathcal{M}_2$ to meshes with pose labels too, in order to be able to select $\mathcal{M}_3$ with the identity label of $\mathcal{M}_1$ and the pose label of $\mathcal{M}_2$. Every 5 epochs, we re-sample a tenth of our training triplets chosen at random. This way, while each triplet is likely to be seen multiple times by the model, which helps lowering the loss values for these specific triplets, the re-sampling allows the model to see new triplets, which helps to better capture the variety of the training set.

Our network takes as input a list of 3D points that correspond to the vertices of the input meshes. We preprocess all meshes by aligning them rigidly and down-sampling them to 2297 vertices using a quadratic error criterion following~\cite{ranjan2018generating}. This down-sampling balances the computing cost of our losses, while keeping a reasonable level of precision. We propose a simple two step up-sampling for better qualitative visualization. First, we up-sample the meshes to 6890 vertices, by placing the new vertices at the centroid of their neighbours~\cite{ranjan2018generating}. Then, we move the new vertices to the local Laplacian coordinates (see Section~\ref{sec:learningobjective}) computed on the unprocessed target identity, while preserving the coordinates of the 2297 vertices predicted by our model.

We use the ADAM optimiser. For the main training, we use a learning rate of $0.001$ and a learning rate decay of $0.99$ per epoch, and train for $500$ epochs. We use batches of size $32$. We set the loss weights in Eq.~\ref{equ:loss} as $\alpha_{rec} = 10$, $\alpha_{lap} = 1000$, $\alpha_{rig} = 1$ and $\alpha_{lat} = 1000$.
For the fine-tuning, we use a learning rate of $0.0001$, and fine-tune for  $50$ iterations. We set the loss weights in Eq.~\ref{equ:loss_ft} as $\alpha_{lap} = 10$, $\alpha_{rig} = 1$ and $\alpha_{reg} = 0.1$.

\section{Evaluation}

In this section we evaluate our model's ability to achieve deformation transfer both quantitatively and qualitatively. We perform an ablation study to evaluate the effects of supervising and fine-tuning our model, and compare our method quantitatively to state-of-the-art deformation transfer methods. In particular, we choose to compare to the supervised NPT~\cite{wang2020neural} and the unsupervised USPD~\cite{zhou20unsupervised} as they achieve the best results in the literature. Finally, we present qualitative results of deformation transfer using our method to extreme identities and characters with simple clothing. We apply our method to animations on a frame-by-frame basis, and to an identity morphing scenario, where the pose stays constant while the identity changes. For additional visualizations, please refer to the supplementary material.

To evaluate the results numerically, we use input pairs of shapes $\mathcal{M}_{p}$ and $\mathcal{M}^{id}$ for which the ground truth transfer $\mathcal{M}_{p}^{id}$ is known. As our meshes are in point-to-point correspondences, the error is measured using the mean of the $L_2$ distances between corresponding vertices of the ground truth $\mathcal{M}_{p}^{id}$ and the result $\tilde{\mathcal{M}}_{p}^{id}$ after Procrustes alignment.

\subsection{Data}
\label{sec:data}



\textbf{ExtFAUST}
To obtain labelled data for supervision, we create a new dataset with full identity and pose labels by augmenting the FAUST dataset~\cite{bogo2014FAUST} with additional pseudo-ground-truth. FAUST contains 10 identities performing the same 10 poses each, providing us with 100 meshes with full identity and pose labels. We extend this data by adding meshes with new poses and identities from other datasets, and then applying an optimization based deformation transfer~\cite{basset2020contact} to transfer every new identity and pose to all pre-existing poses and identities in FAUST. For the new poses and identities added to FAUST, we choose meshes from Dynamic FAUST (DFAUST)~\cite{bogo2017dynamic}, SMPL~\cite{loper2015smpl}, and Mixamo~\cite{mixamo}. We add 11 identities and 17 poses to the original FAUST data, yielding 540 meshes with pose and identity labels after manually removing a few outliers. We refer to the resulting dataset as Extended FAUST (ExtFAUST) in the remainder of this paper. We created a test split by removing all occurrences of  4 poses and 4 identities from this dataset. This leaves 369 shapes for training and 171 for testing.

\textbf{DFAUST}
We also use the Dynamic FAUST dataset (DFAUST)~\cite{bogo2017dynamic} for training. This dataset contains 10 identities performing between 11 and 14 motions, each of them containing a few hundred frames. It comes with identity labels. However, even if the motions are semantically equivalent across subjects, the poses differ in timing and style, and no pose labels are available. The dataset contains 41220 shapes, which are used for training.

\textbf{Test sets}
We use three different test sets in our evaluations. 
The \emph{ExtFAUST pose test set} consists of 4 identities in 4 poses, all of which were unseen during training. This allows to evaluate the method's ability to generalize to both new identities and poses. When combining all possible triplets of target identity, source pose, and transfer ground truth, 240 triplets are available for testing.
The \emph{ExtFAUST id test set} consists of 4 identities unseen during training in 4 poses that were seen during training. This allows to evaluate the method's ability to generalize to identities on poses that it has been trained on. A total of 240 ground truth triplets are available for testing.
The \emph{AMASS test set} is used for evaluation w.r.t. the state-of-the-art. It contains 100 triplets generated from motion capture data used in AMASS~\cite{AMASS:ICCV:2019} combined with random SMPL~\cite{loper2015smpl} shape parameters.

\subsection{Ablation study}
\label{sec:ablation}

To evaluate the necessity and effectiveness of our supervision scheme (Eq.~\ref{equ:loss}), we train our model without any full supervision, with only the FAUST data as full supervision (approximately $0.2\%$ of the training data is labeled), and with all the ExtFAUST data as full supervision (approximately $1\%$ of the training data is labeled). Tab.~\ref{tab:ablation1} reports the errors in $mm$ on the ExtFAUST pose test set. Note that a small percentage of labeled training data allows to improve our results by almost $1cm$.


\begin{table}
    \centering
    \begin{tabular}{|c|c|c|c|}
         \hline
        Supervision & None & FAUST & ExtFAUST \\
         \hline
        Mean error ($mm$) & 29.19 & 24.83 & 20.19 \\
        \hline
    \end{tabular}
    \caption{Ablation study on supervision.}
    \label{tab:ablation1}
\end{table}

To evaluate the influence of the fine-tuning at inference time, we run our method with and without fine-tuning on the ExtFAUST test set. While the error without fine-tuning is $31.51mm$, it decreases significantly to $20.19mm$ when fine-tuning is used.

\subsection{Comparison to state of the art}

\begin{figure*}[ht]
    \centering
    \includegraphics[width=0.75\linewidth]{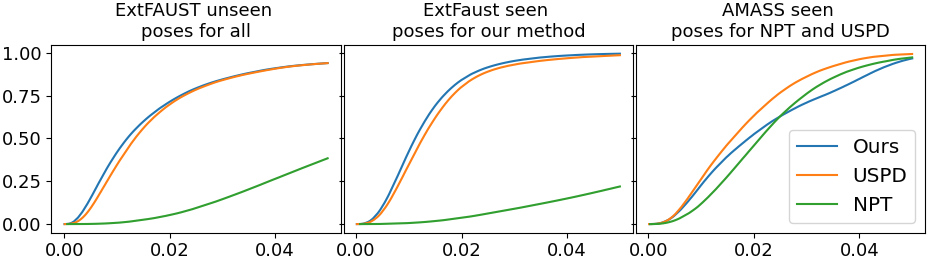}
    \caption{Cumulative errors for our method, USPD and NPT on 3 validation sets. The $x$-axis shows per-vertex errors ($mm$). The $y$-axis is the proportion of all error values below the corresponding error value.}
    \label{fig:cumerror}
\end{figure*}

To the best of our knowledge none of the existing deformation transfer methods operate in a weakly supervised way and we compare therefore our method to the state-of-the-art supervised method NPT~\cite{wang2020neural} and unsupervised method USPD~\cite{zhou20unsupervised}. 
This results in three methods that make different assumptions on their training supervision. Moreover, while our method and USPD assume full correspondence of the 3D input models, NPT is more general and can handle 3D models without correspondence or fixed topology. These differences make a completely fair comparison difficult. To make the comparison as fair as possible, we train each method in its optimal supervision setting, with the training data presented in the original papers.

We evaluate the errors on our three test sets: one that requires pose generalization from all methods, and two that require pose generalization from some of the methods. For the ExtFAUST pose test set, none of the methods have seen during training any of the poses or identities presented at test time. This test set therefore evaluates all method's abilities to generalize to new poses and new identities, and can be considered the hardest test set for all methods. 
For the ExtFAUST identity test set, none of the methods have seen any of the idenitites presented at test time. However, our method has seen the poses, coupled with other identities, during training. This test set therefore requires NPT and USPD to generalize to new poses, while this is not the case for our method.
For the AMASS test set, none of the methods have seen any of the idenitites presented at test time. However, NPT and USPD have seen the poses, coupled with other identities, during training. This test set therefore requires our method to generalize to new poses, while this is not the case for NPT and USPD.



Fig.~\ref{fig:cumerror} shows cumulative error plots for each method on each validation set. Note that our method and USPD obtain significantly better results for the first two validation sets, that require a generalization ability to new poses from NPT. This is because NPT does not use correspondence information, and treats points on the 3D human model that are close-by as neighbors and aims to deform them using similar deformations. In cases where different body parts are close-by or in contact in one input pose but not the other, this creates stretching artifacts that explain the high errors. For the AMASS test set, where NPT does not need to generalize to new poses and the results provide a meaningful measure for NPT's performance, their result is better, but our method still outperforms NPT on average and in the fine details.

\begin{figure}[h]
    \centering
    \includegraphics[width=0.85\linewidth]{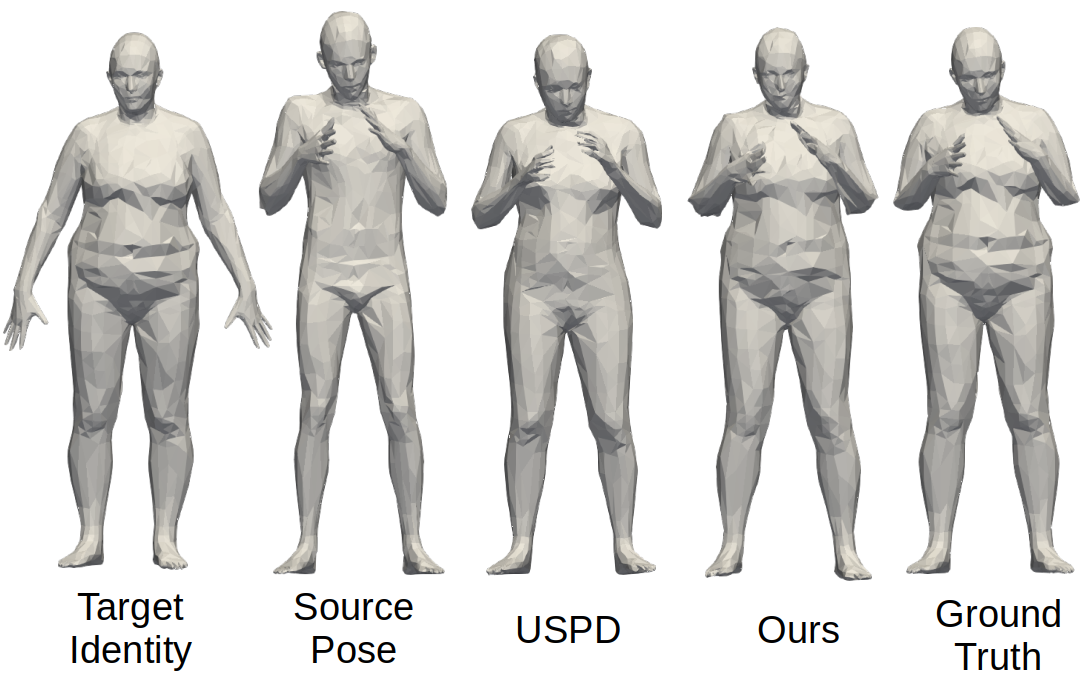}
    \caption{Qualitative comparison to USPD.}
    \label{fig:compUSPD}
\end{figure}

Our method and USPD perform respectively better than the other when one method has seen the poses during training and not the other.
For the ExtFAUST pose test set with poses unseen by all methods, both methods have similar performances, but our method gives slightly better results for the low error range, showing that details are better preserved. 
This can be observed in Fig.~\ref{fig:compUSPD}, where USPD's result has the correct overall body shape, but the details of the identity are overly smooth, whereas our method better transfers the fine-scale geometric details of the identity. It is also noteworthy to mention that USPD uses approximately 3 times more training data than we do.

\subsection{Qualitative evaluation}

\begin{figure}[h]
    \centering
    \includegraphics[width=0.9\linewidth]{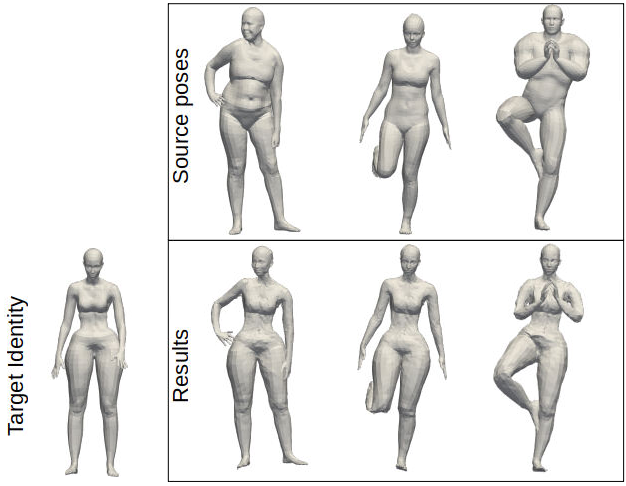}
    \caption{Transferring the identity of an unrealistic character to new poses.}
    \label{fig:extremeshape}
\end{figure}

\begin{figure}[h]
    \centering
    \includegraphics[width=0.9\linewidth]{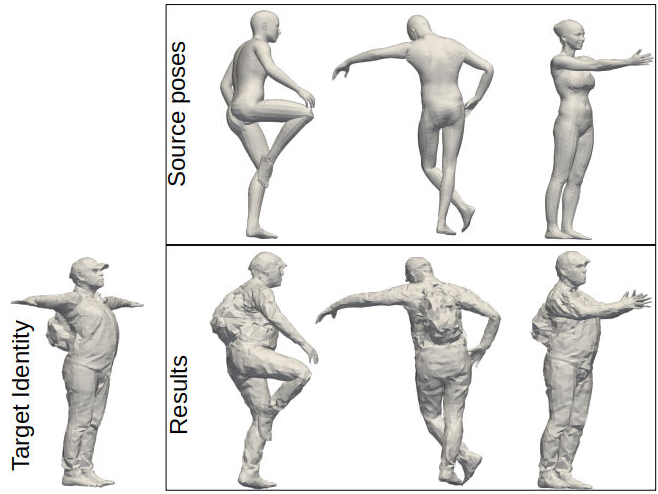}
    \caption{Transferring the identity of a clothed character to new poses.}
    \label{fig:clothes}
\end{figure}

\begin{figure}[h]
    \centering
    \includegraphics[width=0.85\linewidth]{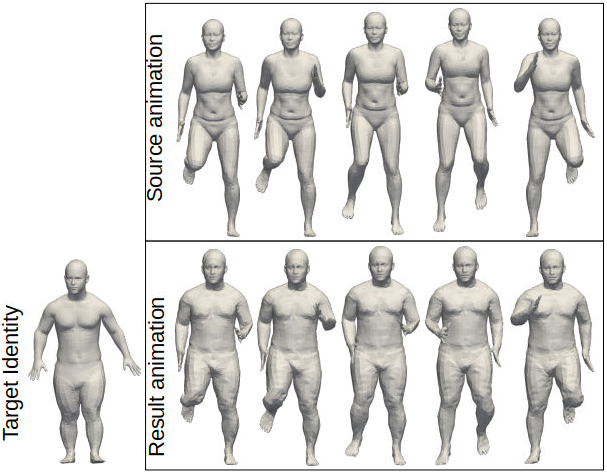}
    \caption{Transferring a new identity to an animation.}
    \label{fig:animation}
\end{figure}

Figures~\ref{fig:extremeshape}, \ref{fig:clothes} and \ref{fig:animation} present results upsampled with the method described in Section~\ref{sec:implementation}, for visualization purposes.

Fig.~\ref{fig:extremeshape} shows results of transferring an unrealistic identity, unseen at training, to new poses. This figure demonstrates that our method is able to transfer extreme identities, while our method only saw realistic identities during training. Fig.~\ref{fig:clothes} shows results of transferring a character with simple clothing and accessories to new poses. Note that during training, no clothed characters or accessories are seen. These results show our method's ability to generalize to data that is far from the distribution of the training data while preserving geometric detail. This property is achieved in large part by the fine-tuning at inference.

To demonstrate the potential of our method, we apply it to two problems arising in automatic content creation.
First, Fig.~\ref{fig:animation} shows our method applied to solve the motion retargeting problem. Given an input animation and a new identity, we apply our method to the animation on a frame-by-frame basis. Note that although no temporal information is used by our method, the resulting animation is consistent and does not suffer from a significant jitter (best observed in supplementary video). 

\begin{figure}[h]
    \centering
    \includegraphics[width=0.8\linewidth]{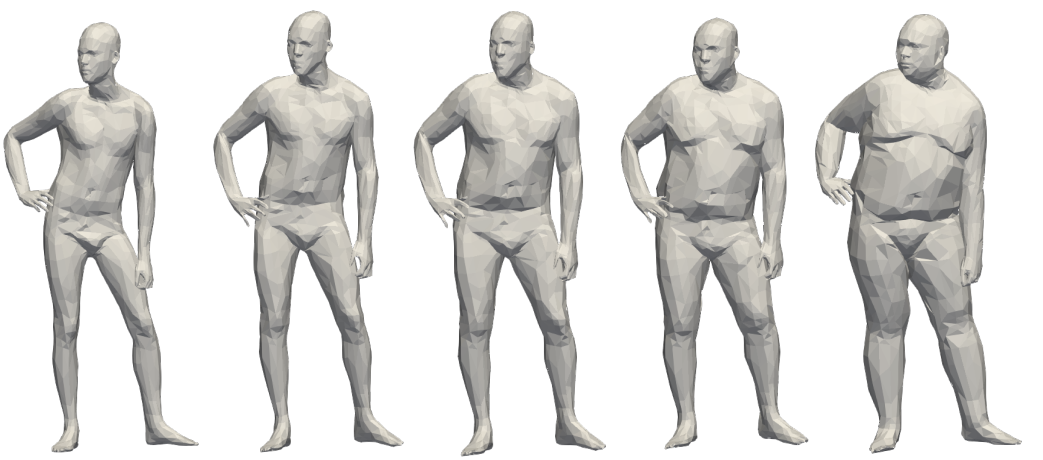}
    \caption{Interpolating the identity latent code between the leftmost and rightmost models.}
    \label{fig:interpolation}
\end{figure}

Second, Fig.~\ref{fig:interpolation} shows our method applied to solve the morphing problem. For this result, the identity code of two input characters  is linearly interpolated in the latent space before being passed to the decoder. Our method is able to interpolate smoothly between identity codes while keeping the pose consistent. In addition to being an interesting application, this result shows that the latent space learned by our method is well structured.

\subsection{Limitations}

While our method gives state-of-the-art results for the deformation transfer problem, limitations remain. First, we require meshes in vertex-to-vertex correspondence. This limits our method to meshes with a  given template, and requires solving a correspondence problem for other inputs.
Another limitation of our method is the need for supervision, as training our network without supervision results in deformed poses due to the near-isometry hypothesis. An interesting direction for future work is to extend our method to be fully unsupervised by exploring other identity losses.

\section{Conclusion}

In this paper, we introduced a neural deformation transfer method that predicts the identity deformation from a source character to a character with the same pose and a new identity. We used geometric properties of meshes to describe identity in a pose invariant way. We introduced a large dataset of human models with full identity and pose labels, which we use in addition to a larger unlabeled dataset to supervise our training. Experiments demonstrate our model's ability to generalize to unseen poses when using around $1\%$ of supervision at training time. A fine tuning step, inspired by the few-shot learning methods, is shown to allow for the transfer of fine-scale geometric details of the identity. The method generalizes well to new identities, and even allows to transfer simple clothing and accessories.


\section*{Acknowledgments}

This project was funded by the Inria IPL AVATAR project, and by the EU’s Horizon 2020 research and innovation program under grant agreement No 952147. We also want to thank Jo\~ao Regateiro for the helpful discussions.

{\small
\bibliographystyle{plain}
\bibliography{main}
}

\end{document}